\documentclass[10pt,twocolumn,letterpaper]{article}

\usepackage[pagenumbers]{cvpr} 

\usepackage{adjustbox}
\usepackage{multirow}
\usepackage{float}
\usepackage{makecell}
\usepackage[most]{tcolorbox}
\usepackage{fontawesome5}

\definecolor{cvprblue}{rgb}{0.21,0.49,0.74}
\usepackage[pagebackref,breaklinks,colorlinks,allcolors=cvprblue]{hyperref}

\def\paperID{22} 
\def\confName{CVPR}
\def\confYear{2026}

\title{Diversity Matters: Dataset Diversification and Dual-Branch Network for Generalized AI-Generated Image Detection}
\author{Nusrat Tasnim\\
Korea Aerospace University\\
10540, Goyang, Gyeonggi, South Korea\\
{\tt\small tasnim.nishu70@kau.kr}
\and
Kutub Uddin\\
University of Michigan-Flint\\
48502, Flint, Michigan, United States\\
{\tt\small kutub@umich.edu}
\and
Khalid Malik\\
University of Michigan-Flint\\
48502, Flint, Michigan, United States\\
{\tt\small drmalik@umich.edu}
}

\begin{document}
\maketitle
\begin{abstract}
The rapid proliferation of AI-generated images, powered by generative adversarial networks (GANs), diffusion models, and other synthesis techniques, has raised serious concerns about misinformation, copyright violations, and digital security. However, detecting such images in a generalized and robust manner remains a major challenge due to the vast diversity of generative models and data distributions. In this work, we present \textbf{Diversity Matters}, a novel framework that emphasizes data diversity and feature domain complementarity for AI-generated image detection. The proposed method introduces a feature-domain similarity filtering mechanism that discards redundant or highly similar samples across both inter-class and intra-class distributions, ensuring a more diverse and representative training set. Furthermore, we propose a dual-branch network that combines CLIP features from the pixel domain and the frequency domain to jointly capture semantic and structural cues, leading to improved generalization against unseen generative models and adversarial conditions.
Extensive experiments on benchmark datasets demonstrate that the proposed approach significantly improves cross-model and cross-dataset performance compared to existing methods. \textbf{Diversity Matters} highlights the critical role of data and feature diversity in building reliable and robust detectors against the rapidly evolving landscape of synthetic content.
\end{abstract}
\vspace{-15pt}    
\section{Introduction}\vspace{-8pt}
\label{sec:intro}

\begin{figure}[t]
    \centering
    {\includegraphics[width=1\linewidth]{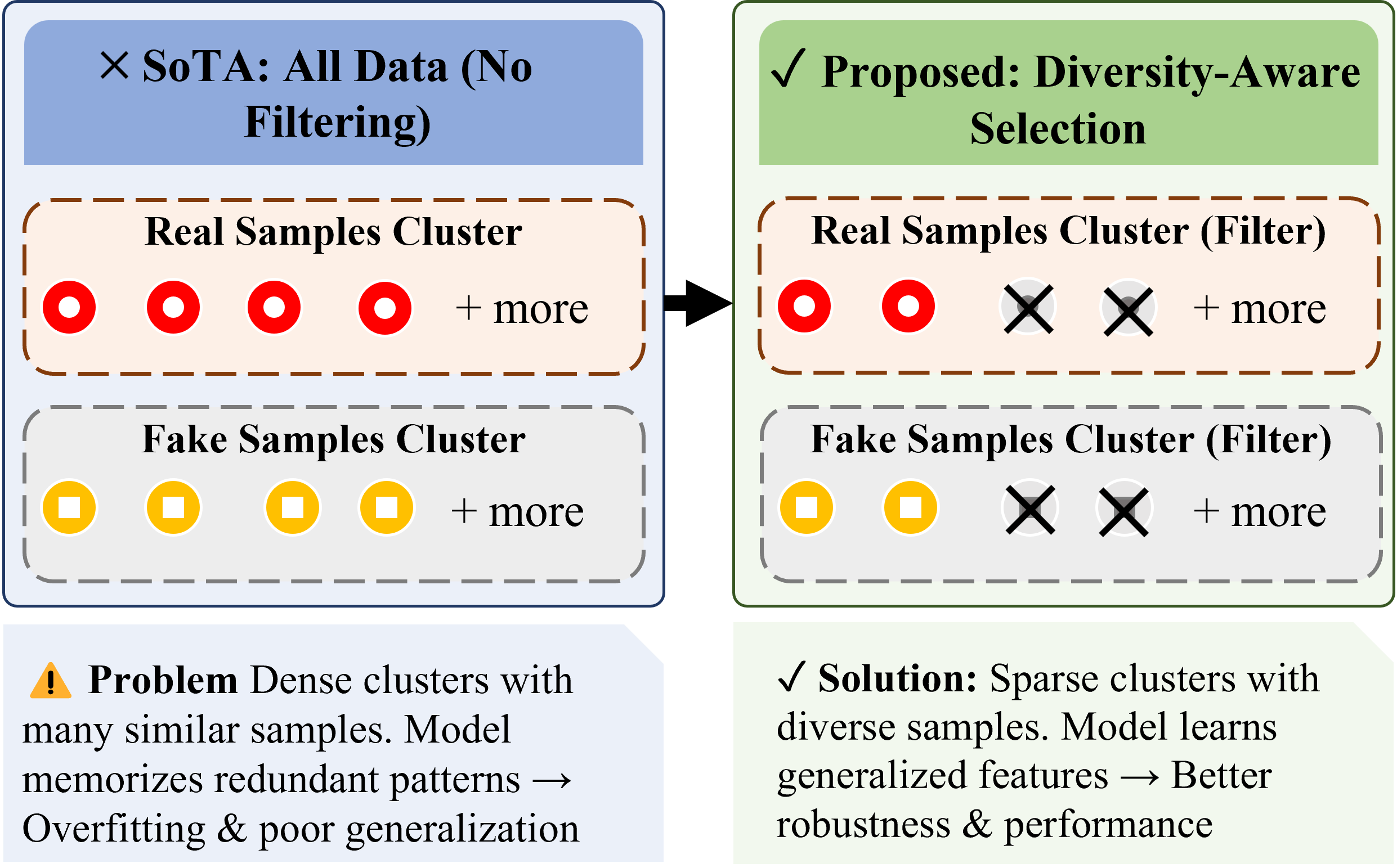}}
    \vspace{-18pt}
    \caption{SoTA vs diversity-aware dataset selection. (Left) Conventional approaches train detectors on all samples with high redundancy, resulting in dense clusters and overfitting. 
    (Right) The proposed CLIP-based diversity-aware approach filters samples across both inter-class and intra-class to retain diverse samples in sparse clusters. Faded samples indicate removed similar pairs. This approach improves the model generalization and robustness while reducing the dataset size significantly for training the detectors.}
    \label{fig:data_diversity}
    \vspace{-25pt}
\end{figure}
In recent years, generative AI has made remarkable progress in synthesizing high-quality images~\cite{MNW2025}, videos~\cite{rossler2019faceforensics++}, and audio~\cite{xie2025codecfake}. Such AI-generated media, particularly images, have rapidly proliferated across social platforms (e.g., Facebook, YouTube, Instagram, TikTok) to spread misinformation. This surge in AI-generated images (AIGIs) poses substantial threats to biometric systems, including face recognition, surveillance, and remote conferencing, and more critically, undermines trust in legal and judicial processes.
For example, in 2024, several AIGIs~\cite{Thete2024} were widely circulated on social media and misinterpreted as real photographs of a flood in Bangladesh. These AIGIs, often depicting emotionally charged scenes, quickly went viral, creating unnecessary public concern and spreading misinformation across the country and beyond. The widespread misinterpretation of such images underscored the growing challenge of distinguishing authentic content from AI-generated fabrications, especially during emergencies such as natural disasters.\\
Recent studies reveal that AIGIs are increasingly being used for fraud, misinformation, and identity attacks. For example, a large-scale analysis of Twitter (X) profiles found that more than 7,700 accounts were using AI-generated faces for spam and influence operations~\cite{TwitterAIGANStudy2024}. During the 2024 U.S. election, approximately 12\% of images shared in political discussion spaces were AIGI, with a small group responsible for nearly 80\% of fake content~\cite{ElectionAIGen2024}. Financial losses are also increasing. AI-enabled fraud caused over \$200 million in damages in early 2025 alone~\cite{DeloitteDeepfake2025}. In addition, surveys show that 68\% of people are concerned about misinformation from AI-generated media, yet less than 1\% can reliably distinguish real from AIGIs~\cite{SecurityMagDeepfake2024}. These statistics highlight the urgent need for generalized and robust AIGI detection and policy interventions to mitigate AI-enabled fraud.\\
In recent years, various AIGI detection methods~\cite{tan2023learning, uddin2023robust, tan2024rethinking, tan2024frequency, yan2024sanity, uddin2024counter, corvi2023detection, tasnim2025ai, tasnim2026grex} have been proposed, following either traditional deep learning paradigms or foundation model–based strategies. Traditional approaches typically involve training models using traditional deep learning as a backbone~\cite{tan2023learning, tan2024rethinking, tan2024frequency}. Foundation model–based approaches~\cite{ojha2023towards, cozzolino2024raising, koutlis2024leveraging, c2pclip2025}, in contrast, exploit large-scale pre-trained encoders, such as CLIP~\cite{radford2021clip}, which can capture diverse semantic and low-level features. A few methods~\cite{frank2020leveraging, qian2020thinking, liu2024forgery, tan2024frequency} used frequency information for detecting synthetic natures. These models provide better generalization to seen generative techniques or similar signatures, including GANs and diffusion models. Nevertheless, these methods face the fundamental challenge to completely new synthetic artifacts, highlighting the persistent difficulty in achieving truly generalized detection, and raising the following research question (RQ):
\vspace{-5pt}
\begin{tcolorbox}[title={\textcolor{yellow}{\faExclamationTriangle}\ Are existing methods sufficiently generalizable to unseen generative models?}]\vspace{-5pt}
\textbf{Answer:} Achieving generalization to unseen or newly developed generative models remains highly challenging. Models trained solely on GAN-based datasets typically fail to generalize to diffusion models, and vice versa. Moreover, relying heavily on large volumes of data from a single category (e.g., GAN or diffusion) often leads to overfitting, which further limits the model's ability to detect previously unseen manipulations.\vspace{-5pt}
\end{tcolorbox}
\vspace{-5pt}
Existing approaches primarily rely on either pixel-domain~\cite{wang2020cnn, barni2020cnn, rajan2024aligned, rajan2025stay, ojha2023towards} or frequency-domain~\cite{tan2023learning, tan2024frequency} information for generalization. Some recent methods~\cite{liu2024forgery, chen2025forgelensdataefficientforgeryfocus} incorporate frequency modules into pretrained models, such as CLIP, to capture spectral artifacts. However, these approaches remain inherently single-branch or loosely coupled, limiting their ability to jointly model semantic and structural inconsistencies.\\
To overcome the limitations of existing approaches~\cite{tan2023learning, tan2024rethinking, ojha2023towards, cozzolino2024raising, liu2024forgery, tan2024frequency}, we introduce a novel AIGI detection framework that (i) diversifies the training data inspired by Abbas et al.~\cite{abbas2023semdedup}, including both GAN- and diffusion-based generative models by incorporating feature-space filtering, and (ii) employs a dual-branch architecture designed to enhance generalization to previously unseen generators, as illustrated in Figure~\ref{fig:data_diversity} (right). 
The proposed dual-branch design explicitly integrates complementary pixel- and frequency-domain representations, enabling more comprehensive artifact modeling and significantly improving generalization across unseen generative models.
The key contributions are summarized as follows:
\begin{itemize}[noitemsep,topsep=2pt]
\item We found that training with a single generative model, such as ProGAN~\cite{karras2017progressive}, is insufficient for generalization to unseen samples, particularly new diffusion models.
\item We introduce dataset diversification via feature-space filtering with a pretrained encoder, removing redundant data during training to capture diverse characteristics.
\item We propose a dual-branch network that combines pixel-domain and frequency-domain information, enhancing overall generalization.
\item Extensive experiments across multiple AIGI detection models and benchmark datasets demonstrate the effectiveness of our approach.
\end{itemize}

\section{Related Works}\vspace{-8pt}
This section presents an overview of existing AIGI detection methods, spanning from traditional deep learning approaches to foundation model–based strategies.\vspace{-5pt}
\label{sec:related}
\subsection{Traditional Deep Learning-Based Approaches}\vspace{-4pt}
Early AIGI detection methods relied heavily on traditional deep learning architectures, especially convolutional neural networks (CNNs) such as ResNet~\cite{he2016deep}, InceptionNet~\cite{szegedy2016rethinking}, XceptionNet~\cite{chollet2017xception}, DenseNet~\cite{huang2017densely}. These methods exploited spatial artifacts, texture inconsistencies, and blending irregularities that emerged during manipulation. For example, Wang et al~\cite{wang2020cnn} trained a ResNet18~\cite{he2016deep} model with the ProGAN~\cite{karras2017progressive} dataset by applying careful augmentation and showed promising generalization across unseen samples. After this method, most methods followed the same protocol to train the detection model on the ProGAN dataset and generalize across unseen samples. Tan et al.~\cite{tan2023learning} introduced a gradient-based approach in which they used the gradient of pre-trained models. Corvi et al.~\cite{corvi2023detection} explored the diffusion artifacts and used frequency information to detect the synthetic signature. Tan et al.~\cite{tan2024rethinking} explored the upsampling properties in AIGIs and formalized neighboring pixel relations (NPR) for generalized detection. They further developed a frequency-aware network to capture low and high frequency cues for AIGIs detection. Rajan et al.~\cite{rajan2024aligned} argued that a better-aligned real and AIGI dataset helped in improving generalization across diffusion models. Later, they proposed a stay positive method, in which they first trained a detector on both real and AIGI samples, and then retrained the model using only AIGIs to enhance its focus on synthetic artifacts. 
\vspace{-5pt}
\subsection{Foundation Model-Based Approaches}\vspace{-4pt}
With the rise of large-scale pre-trained models, foundation models, particularly contrastive models such as CLIP, have emerged as powerful tools for AIGI detection. Several methods~\cite{ojha2023towards, cozzolino2024raising, liu2024forgery, chen2025forgelensdataefficientforgeryfocus} leverage these models either as feature extractors or by fine-tuning them on targeted datasets. 
For example, Ojha et al.~\cite{ojha2023towards} used CLIP as a frozen feature extractor and trained a detection head to identify AIGIs. Cozzolino et al.~\cite{cozzolino2024raising} evaluated CLIP with a small number of samples and achieved promising performance. Koutlis et al.~\cite{koutlis2024leveraging} extracted class token features from an intermediate CLIP encoder and introduced a trainable importance estimator to improve generalization across unseen samples. 
Tan et al.~\cite{c2pclip2025} generated a common caption for all real and fake samples and injected it into the encoder to enhance generalization. Liu et al.~\cite{liu2024forgery} proposed a forgery-aware adapter to fine-tune CLIP on AIGI datasets. Similarly, Yan et al.~\cite{yan2024effort} applied a singular value decomposition technique to fine-tune CLIP for better performance on AIGI datasets. Chen et al.~\cite{chen2025forgelensdataefficientforgeryfocus} further extended CLIP by fine-tuning it with forgery-specific features using a weight-shared guidance module, achieving improved generalization.\vspace{-6pt}
\subsection{Limitation of the Existing Methods}\vspace{-5pt}
A major limitation of existing methods~\cite{wang2020cnn, tan2023learning, corvi2023detection, ojha2023towards, liu2024forgery, chen2025forgelensdataefficientforgeryfocus} is that most focus on either GAN- or diffusion-based generative models and are evaluated on a limited set of datasets. Moreover, most approaches~\cite{wang2020cnn, corvi2023detection, liu2024forgery} train their models using only one generative model, which restricts the decision boundary and can lead to overfitting. As new generative models continue to emerge, these methods often fail to generalize, since they are tailored to specific datasets. In addition, relying on large volumes of data from a single generative model further limits their ability to generalize to unseen synthetic content. \vspace{-4pt}
\section{Proposed Method}\vspace{-4pt}
\label{sec:method}
This section details the proposed method, outlining its motivation, defining dataset diversity, and describing the overall AIGI detection framework.\vspace{-4pt}
\subsection{Problem Statement and Motivation}\vspace{-4pt}
AIGI detection is increasingly difficult as generative models, spanning GANs, diffusion models, and perceptual synthesis pipelines, grow more diverse and realistic. Each generative approach introduces unique architectural priors and artifact signatures, making it impractical for a single detector to capture all synthetic cues. Concretely, models trained exclusively on GAN datasets (e.g., ProGAN) fail to generalize to diffusion, and vice versa. In addition, most existing detectors are trained on large volumes of data from a single generative source, causing them to overfit to narrow, model-specific feature distributions rather than learning broadly applicable synthetic cues.\\
These limitations manifest in two practical failure modes: (i) data redundancy, where densely clustered training samples reinforce narrow decision boundaries, and (ii) feature incompleteness, where relying solely on pixel-domain representations misses the structural and spectral artifacts that are more clearly exposed in the frequency domain. Addressing both failure modes is essential: a diverse, deduplicated training set reduces model-specific bias, while complementary pixel- and frequency-domain features together capture the full range of synthetic signatures needed for robust, cross-domain detection.

\begin{figure}[t]
    \centering
    {\includegraphics[width=0.9\linewidth]{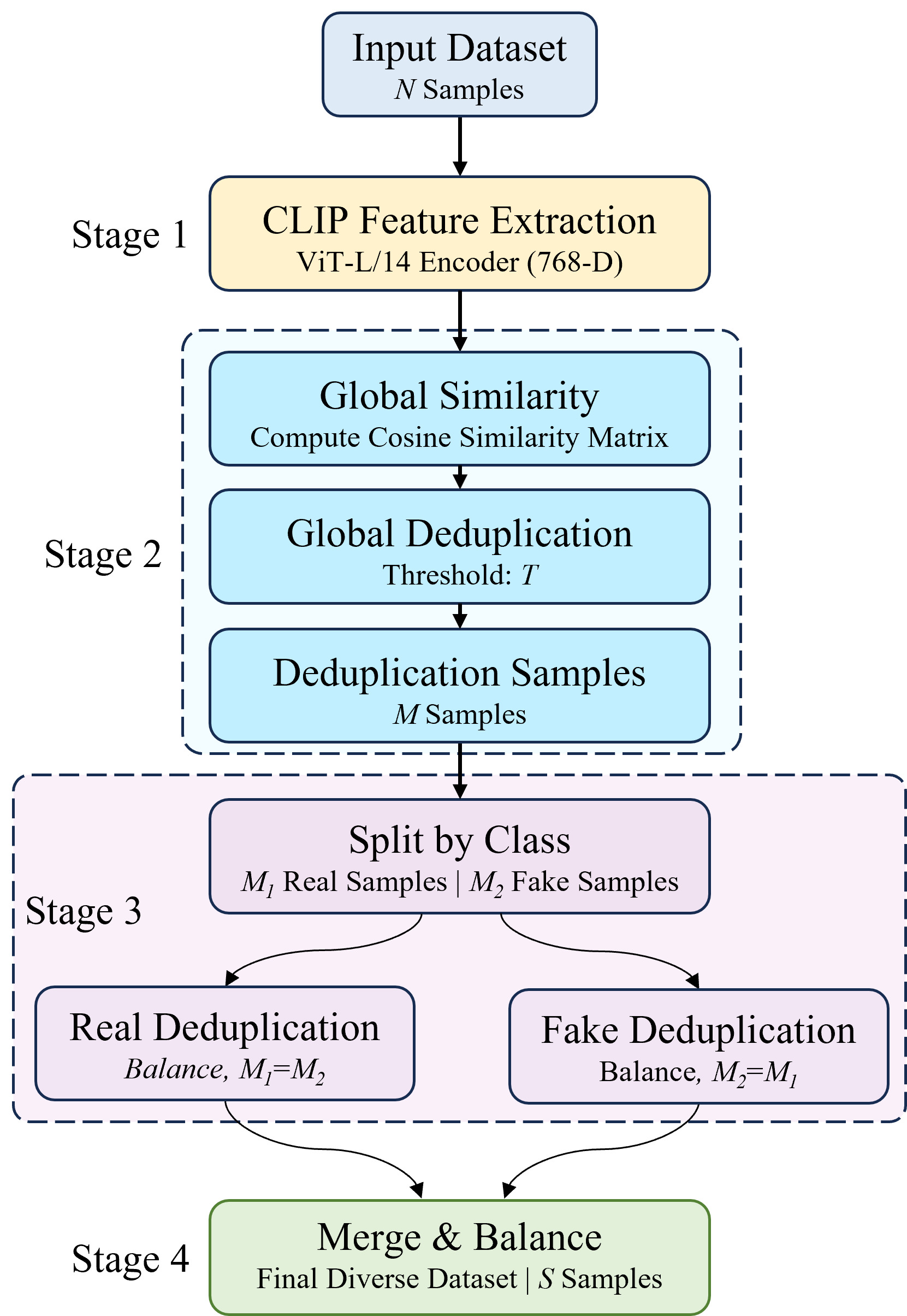}}
    \vspace{-5pt}
    \caption{Flow diagram of the proposed data diversification to improve generalizability across diverse generative models.}
    \vspace{-18pt}
    \label{fig:diversity}
\end{figure}

\begin{figure*}[t]
    \centering
    {\includegraphics[width=1\linewidth]{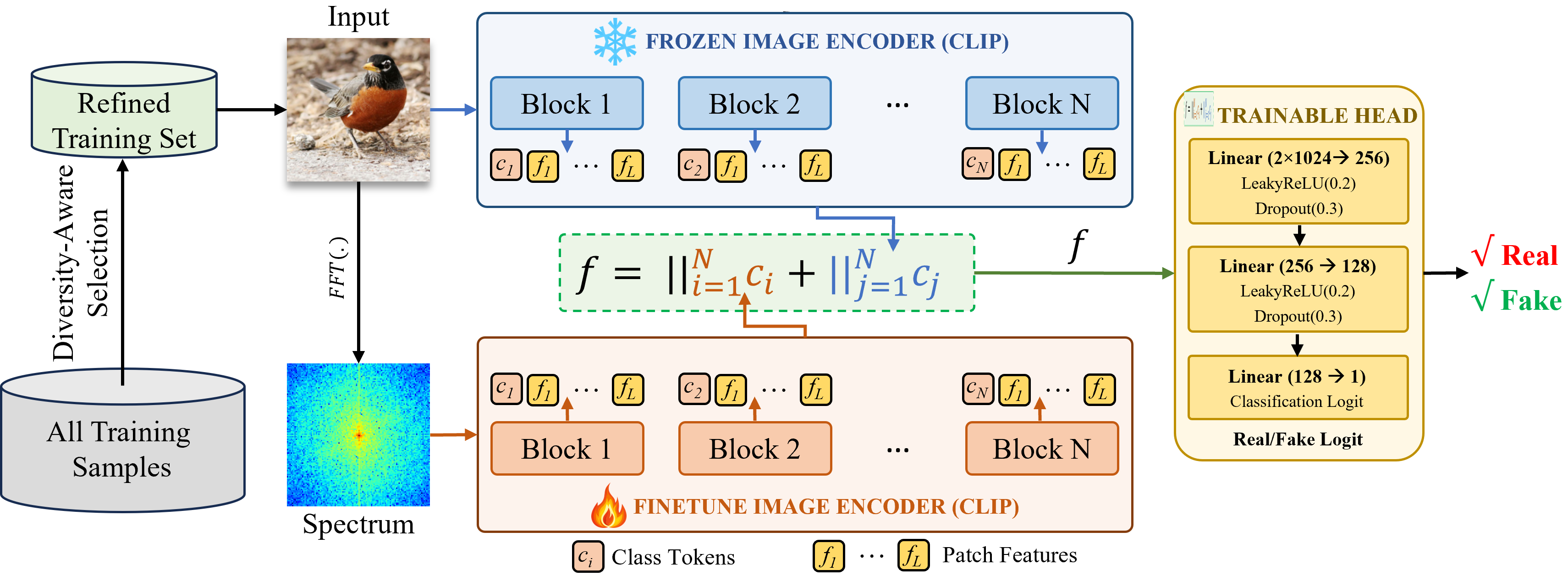}}
    \vspace{-18pt}
    \caption{Architecture of the proposed diversity-aware and dual-branch AIGI detection framework. A diversity-aware selection strategy curates a representative training set. 
    In the first branch (top), a frozen CLIP image encoder extracts embeddings from each patch of the real pixel-level input. 
    In the second branch, the CLIP image encoder is fine-tuned using the corresponding frequency-spectrum patches. 
    Class tokens from all patches across both branches are then concatenated and passed through a trainable detection head. 
    The detection head predicts whether the input sample is real or fake.}
    \vspace{-15pt}
    \label{fig:arch}
\end{figure*}

\subsection{Formulation of Dataset Diversity}\vspace{-4pt}
AIGI datasets often contain high redundancy, where many samples exhibit near-identical artifacts or visual patterns. Training on such data can bias the detector toward narrow feature distributions. To address this limitation, we introduce a feature-domain similarity filtering strategy to discard redundant samples, thereby bringing diversity and improving generalization. Figure~\ref{fig:diversity} illustrates a four-stage deduplication pipeline designed to create a balanced, diverse dataset while removing duplicates:\\
\textbf{Stage 1 -- Feature Extraction:} The input dataset of $N$ samples is processed through a CLIP ViT-L/14~\cite{radford2021learning} encoder (768-D) to extract visual features, enabling efficient similarity comparison for deduplication.\\
\textbf{Stage 2 -- Global Deduplication:} A cosine similarity is computed to identify globally similar samples, which are then deduplicated using a threshold $T$, reducing the dataset to $M$ samples. We employ cosine similarity because CLIP embeddings are optimized for directional alignment in the feature space. It focuses on the angular relationship between feature vectors, capturing semantic similarity while being invariant to feature magnitude.\\
\textbf{Stage 3 -- Class-Based Threshold Refinement:} The deduplicated samples are split into real and fake classes ($M_1$ and $M_2$ respectively). The pipeline then applies deduplication based on class, on the condition $M_1 > M_2$ to make the real and fake classes balance.\\
\textbf{Stage 4 -- Final Balancing:} The separately deduplicated real and fake samples are merged to produce the final diverse dataset of $S$ samples, which maintains class balance and removes redundancy. \\
Let $\mathcal{X} = \{x_i\}_{i=1}^N$ denote the set of real and fake images obtained from multiple generative sources (most commonly used GAN and diffusion techniques). For each sample $x_i$, we extract its CLIP-based representation, defined as follows:
\vspace{-3pt}
\begin{equation}
    f_i = \phi(x_i)
\end{equation}
where $\phi(\cdot)$ is a pre-trained CLIP encoder. We compute pairwise cosine similarity ($\mathcal PCS$) and discard samples whose similarity exceeds a threshold $T$:
\vspace{-5pt}
\begin{equation}
\begin{aligned}
    \mathcal{PCS}(f_i, f_j) &= 
    \frac{f_i^\top f_j}{\|f_i\|\,\|f_j\|} \\
    \text{discard}(x_j)\quad &\text{if}\quad \text{PCS}(f_i, f_j) > T
\end{aligned}
\end{equation}
This filtering ensures that the retained subset
\vspace{-5pt}
\begin{equation}
    \mathcal{X}^\ast \subseteq \mathcal{X}
    \vspace{-8pt}
\end{equation}
is more diverse and representative across different generative models. By constructing a dataset with reduced redundancy, the detector is encouraged to learn broader, more generalizable cues associated with synthetic content.

\subsection{Architecture of the Proposed AIGI Detector}\vspace{-4pt}
Figure~\ref{fig:arch} depicts the overall architecture of the proposed detector. The architecture employs a dual-branch design that processes complementary representations of the input: (i) the pixel domain (raw image) and (ii) the frequency domain (spectral). The first branch utilizes a frozen image encoder with pretrained weights, while the second branch uses a finetuned image encoder with the same architecture. Both branches extract class tokens and patch features, in which only class tokens are aggregated through concatenation to form a combined feature representation. This fused feature representation is then fed into a trainable detection head consisting of three linear layers with LeakyReLU activation and dropout regularization, progressively reducing the feature dimensionality from 2$\times$1024 to 256 to 128, followed by a final linear layer that produces a binary logit for distinguishing between real and fake images.

\subsubsection{Dual-Branch Network}\vspace{-4pt}
Given an input image $x$, we define two parallel pathways:
\textbf{Pixel-Domain Branch:} 
This branch uses CLIP's visual encoder to extract a semantic and appearance-based representation ($R_p$):
\vspace{-5pt}
\begin{equation}
    R_p = \phi_{\text{pixel}}(x)
    \vspace{-3pt}
\end{equation}
These features capture semantic inconsistencies and unnatural visual patterns produced by generative models.\\
\textbf{Frequency-Domain Branch.} AIGIs often contain structural irregularities and periodic patterns that are more easily revealed in the frequency spectrum than in the spatial domain. To exploit these artifacts, we first transform the input image into the frequency domain using the discrete Fourier transform ($\mathcal DFT$)~\cite{durall2020watch}, which decomposes the image into its spectral components. The resulting magnitude spectrum highlights subtle generative inconsistencies such as repetitive textures, patch-level harmonics, and abnormal high-frequency attenuation. This frequency representation is then fed into a CLIP-based encoder to extract high-level spectral features that complement the semantic cues learned in the pixel domain.
The frequency-domain processing begins with the DFT transformation:
\vspace{-3pt}
\begin{equation}
    x_s = \mathcal {DFT}(x)
    \vspace{-6pt}
\end{equation}
where $x$ is the input image and $x_f$ is its frequency spectrum. The magnitude spectrum is subsequently processed by a finetuned CLIP-based encoder to extract spectral representation ($R_s$):
\vspace{-3pt}
\begin{equation}
    R_s = \phi_{spectrum}(x_s)
\end{equation}
where $\phi_{\text{spectrum}}$ represents the finetuned encoder. The feature representations from both the pixel-domain branch $R_p$ and spectral-domain branch $R_s$ are concatenated at the token level to leverage complementary information:
\vspace{-3pt}
\begin{equation}
    f = [f_p \,\|\, f_s]
    \vspace{-6pt}
\end{equation}
where the pixel-domain branch features are aggregated as:
\vspace{-3pt}
\begin{equation}
    f_p = \|_{i=1}^{N} R_{p_i}
    \vspace{-6pt}
\end{equation}
representing the concatenation of class tokens from the $N$ blocks of the spatial encoder, , $R_p$, and the frequency-domain branch features are aggregated as:
\vspace{-3pt}
\begin{equation}
    f_s = \|_{j=1}^{N} R_{s_i}
    \vspace{-6pt}
\end{equation}
representing the concatenation of class tokens from the $N$ blocks of the spectral encoder, $R_s$. The final fused feature representation $f$ combines both domain-specific features through concatenation and is subsequently fed into the classification head for the real/fake prediction.

\subsubsection{Loss Function}\vspace{-4pt}
We optimize the proposed dual-branch detector using a hybrid objective that combines cross-entropy classification loss with supervised contrastive~\cite{koutlis2024leveraging} regularization:
\vspace{-3pt}
\begin{equation}
    \mathcal{L} = \mathcal{L}_{\text{CE}}(y, \hat{y}) 
    + \lambda \mathcal{L}_{\text{SC}}(y, f)
\end{equation}
where $\mathcal{L}_{\text{CE}}$ enforces correct real/fake classification, $\mathcal{L}_{\text{SC}}$ promotes representation complementarity between the two branches, and $\lambda$ controls the regularization strength.
The cross-entropy loss is defined as:
\vspace{-3pt}
\begin{equation}
    \mathcal{L}_{\text{CE}}(y, \hat{y}) = -\left[ y \log(\hat{y}) + (1-y) \log(1-\hat{y}) \right]
\end{equation}
where $y \in \{0, 1\}$ is the ground truth label and $\hat{y}$ is the predicted probability.\\
The supervised contrastive loss ($\mathcal{L}_{\text{SC}}$) regularizes the feature space by pulling together representations from the same class while pushing apart different classes, ensuring both branches learn complementary and discriminative features that improve generalization, defined as follows:
\vspace{-3pt}
\begin{equation}
\mathcal{L_{SC}} = \frac{1}{B} \sum_{i=1}^{B} 
\frac{-1}{|P(i)|} \sum_{p \in P(i)} 
\log \frac{\exp\left( \frac{z_i \cdot z_p}{\tau} \right)}
         {\sum_{a \neq i} \exp\left( \frac{z_i \cdot z_a}{\tau} \right)}
 \vspace{-6pt}
\end{equation}
where $B$ is batch size, $z_i$ and $z_p$ are features of anchor and positives, $P(i)$ is positive set, and $z_a$ are other samples.

\section{Results}
\label{results}
\vspace{-8pt}
This section presents the experimental setup, performance evaluation, and comparisons, as well as a discussion.
\vspace{-5pt}
\begin{table}[!t]
\centering
\caption{Benchmark AI-Generated Datasets: Overview of datasets used in this study, categorized by generative paradigm, with release year, generative techniques, and sample counts.} 
\vspace{-8pt}
\begin{adjustbox}{width=1\linewidth}
\begin{tabular}{lcccc}
\hline
\textbf{\makecell[l]{Name of \\ Dataset}}& \textbf{\makecell[c]{Year of \\ Release}} & \textbf{\makecell[c]{Generative \\ Techniques}} & \textbf{\makecell[c]{Real \\ Samples}} & \textbf{\makecell[c]{Generated \\ Samples}} \\
\hline
\multicolumn{5}{l}{\textbf{GAN-based Datasets}} \\ 
\hline
PGAN~\cite{karras2017progressive} & 2020 & 1 & 72,012 & 72,012 \\
CNNDF~\cite{wang2020cnn} & 2020 & 8 & 31,006 & 30,997\\
\hline
GANDF~\cite{tan2024rethinking} & 2024 & 9 & 18,000 & 18,000\\
\hline
\multicolumn{5}{l}{\textbf{Diffusion-based Datasets}} \\
\hline
D1KS~\cite{tan2024rethinking} & 2023 & 5 & 18,000 & 17,992\\
\hline
UDF~\cite{ojha2023towards} & 2023 & 8 & 8,000 & 8,000\\
\hline
DIRE~\cite{wang2023dire} & 2024 & 8 & 18,000 & 22,603\\
\hline
GENI~\cite{zhu2023genimage} & 2023 & 9 & 46,999 & 47,000\\
\hline
MNW~\cite{MNW2025} & 2025 & 43 & 0 & 11,250\\
\hline
\multicolumn{5}{l}{\textbf{Mixed Dataset}} \\
\hline
MGD~\cite{wang2020cnn, ojha2023towards} & - & 20 & 52,169 & 52,160\\
\hline
\end{tabular}
\end{adjustbox}
\label{tab:dataset_summary}
\vspace{-18pt}
\end{table}

\begin{table*}[!t]
\centering
\caption{Performance of different feature combinations across backbone networks for various numbers of training samples. The same encoders are used for dataset diversification and the detection pipeline. The proposed model is trained using samples from PGAN~\cite{karras2017progressive} and D1KS~\cite{abdulrahman2025deepfake}, while the evaluation is performed on the MGD~\cite{wang2020cnn} dataset to measure its generalization performance.}
\vspace{-8pt}
\begin{adjustbox}{max width=\textwidth}
\begin{tabular}{c|c|c|ccc|ccc|ccc|ccc|ccc|ccc}
\toprule
\midrule
\multirow{2}{*}{\textbf{Model}} & \multicolumn{2}{c|}{\textbf{Features}} 
& \multicolumn{3}{c|}{\textbf{4K Samples, T=0.6}} 
& \multicolumn{3}{c|}{\textbf{9K samples, T=0.7}} 
& \multicolumn{3}{c|}{\textbf{19K samples, T=0.8}} 
& \multicolumn{3}{c|}{\textbf{42K samples, T=0.85}} 
& \multicolumn{3}{c|}{\textbf{97K samples, T=0.9}} 
& \multicolumn{3}{c}{\textbf{180K samples, T=1.0}} \\
\cmidrule(lr){2-3} \cmidrule(lr){4-6} \cmidrule(lr){7-9} \cmidrule(lr){10-12} 
\cmidrule(lr){13-15} \cmidrule(lr){16-18} \cmidrule(lr){19-21}
 & $f_p$ & $f_s$ & ACC & AUC & AP & ACC & AUC & AP & ACC & AUC & AP & ACC & AUC & AP & ACC & AUC & AP & ACC & AUC & AP \\
\midrule
    \multirow{3}{*}{VGG-16~\cite{simonyan2014very}}
    & \checkmark & $\times$ & 50.0 & 59.7 & 57.8 & 50.0 & 41.1 & 45.3 & 50.0 & 56.4 & 54.3 & 50.0 & 60.1 & 58.5 & 50.0 & 50.0 & 50.0 & 50.0 & 50.0 & 50.0 \\
    & $\times$ & \checkmark & 50.0 & 37.6 & 42.7 & 50.0 & 61.3 & 61.6 & 50.0 & 53.6 & 51.9 & 50.0 & 50.0 & 50.0 & 50.0 & 50.0 & 50.0 & 50.0 & 50.0 & 50.0 \\
    & \checkmark & \checkmark & 50.0 & 47.6 & 48.8 & 50.0 & 41.4 & 45.6 & 50.0 & 50.0 & 50.0 & 50.0 & 50.0 & 50.0 & 50.0 & 50.0 & 50.0 & 50.0 & 50.0 & 50.0 \\
\midrule
\multirow{3}{*}{ResNet-18~\cite{he2016deep}}
    & \checkmark & $\times$ & 50.0 & 54.1 & 52.5 & 50.0 & 53.6 & 53.5 & 47.6 & 49.2 & 47.8 & 50.8 & 50.9 & 50.5 & 50.0 & 51.4 & 53.8 & 50.0 & 50.0 & 50.0 \\
    & $\times$ & \checkmark & 56.0 & 54.4 & 50.5 & 51.9 & 51.4 & 52.4 & 52.8 & 52.2 & 49.2 & 50.0 & 42.6 & 46.7 & 50.0 & 50.0 & 50.0 & 50.0 & 50.0 & 50.0 \\
    & \checkmark & \checkmark & 53.3 & 61.7 & 63.5 & 57.0 & 62.6 & 63.8 & 55.9 & 57.9 & 54.1 & 53.3 & 54.9 & 53.8 & 50.0 & 50.0 & 50.0 & 50.0 & 50.0 & 50.0 \\
\midrule
\multirow{3}{*}{MobileNet-V3~\cite{howard2019searching}}
    & \checkmark & $\times$ & 60.0 & 64.2 & 63.0 & 69.9 & 77.3 & 76.8 & 69.5 & 75.5 & 75.9 & 72.7 & 81.3 & 82.0 & 74.4 & 82.5 & 82.9 & 73.7 & 81.6 & 83.9 \\
    & $\times$ & \checkmark & 55.0 & 55.5 & 53.6 & 56.6 & 57.8 & 55.1 & 57.7 & 61.2 & 59.6 & 60.4 & 62.5 & 58.3 & 59.4 & 61.7 & 59.0 & 59.6 & 62.3 & 61.1 \\
    & \checkmark & \checkmark &  67.7 & 71.6 & 73.5 & 76.1 & 82.9 & 84.7 & 78.3 & 87.3 & 86.4 & 79.7 & 88.0 & 87.4 & 79.2 & 87.9 & 88.7 & 78.6 & 86.0 & 82.9 \\
\midrule
\multirow{3}{*}{EfficientNet-B4~\cite{tan2019efficientnet}}
    & \checkmark & $\times$ & 67.4 & 74.4 & 75.3 & 70.2 & 78.6 & 78.9 & 75.4 & 83.8 & 83.5 & 81.2 & 87.0 & 86.8 & 78.9 & 85.2 & 86.2 & 80.3 & 89.4 & 90.8 \\
    & $\times$ & \checkmark & 57.6 & 59.3 & 55.3 & 57.8 & 60.8 & 57.8 & 65.3 & 68.3 & 65.5 & 50.1 & 48.7 & 49.4 & 63.4 & 69.5 & 68.2 & 51.3 & 51.3 & 50.8 \\
    & \checkmark & \checkmark & 70.6 & 77.7 & 78.1 & 76.9 & 85.3 & 84.7 & 72.1 & 83.2 & 76.3 & 64.9 & 85.3 & 84.2 & 81.9 & 87.2 & 81.2 & 78.9 & 83.0 & 83.4 \\
\midrule
\midrule
    \multirow{3}{*}{Dinov2-Base~\cite{oquab2023dinov2}}
    & \checkmark & $\times$ & 75.9 & 87.2 & 88.5 & 78.9 & 90.1 & 91.2 & 78.9 & 87.6 & 88.0 & 78.8 & 84.7 & 81.8 & 78.6 & 89.2 & 90.0 & 79.5 & 91.1 & 92.3 \\
    & $\times$ & \checkmark & 61.1 & 66.7 & 65.5 & 63.1 & 67.3 & 62.3 & 65.2 & 71.0 & 66.6 & 70.2 & 77.6 & 73.9 & 70.5 & 79.0 & 76.5 & 71.9 & 80.0 & 78.3 \\
    & \checkmark & \checkmark & 76.7 & 88.1 & 89.4 & 78.5 & 89.2 & 90.0 & 79.3 & 84.3 & 82.6 & 78.6 & 87.0 & 86.7 & 78.8 & 88.1 & 88.8 & 77.8 & 87.5 & 88.7 \\
\midrule
\multirow{3}{*}{Dinov2-Large~\cite{oquab2023dinov2}}
    & \checkmark & $\times$ & 79.9 & 91.6 & 92.4 & 80.5 & 92.0 & 92.6 & 81.4 & 90.5 & 90.4 & 81.8 & 92.7 & 92.9 & 81.2 & 92.1 & 92.4 & 81.1 & 91.9 & 92.1 \\
    & $\times$ & \checkmark & 68.7 & 76.2 & 73.4 & 70.5 & 79.6 & 75.8 & 74.2 & 82.6 & 80.2 & 75.7 & 84.6 & 82.9 & 76.1 & 84.8 & 82.4 & 77.3 & 86.3 & 86.0 \\
    & \checkmark & \checkmark & 79.2 & 91.5 & 92.6 & 81.5 & 93.4 & 93.5 & 81.2 & 90.7 & 90.2 & 82.2 & 91.4 & 90.6 & 81.7 & 93.5 & 94.2 & 81.9 & 91.8 & 92.4 \\
\midrule
\midrule
    \multirow{3}{*}{ViT-B/16~\cite{radford2021learning}}
    & \checkmark & $\times$ & 81.0 & 89.2 & 88.7 & 82.6 & 92.4 & 92.2 & 82.9 & 90.4 & 89.8 & 83.7 & 92.0 & 91.8 & 83.8 & 94.7 & 95.0 & 82.6 & 92.2 & 92.0 \\
    & $\times$ & \checkmark & 66.5 & 74.0 & 70.4 & 69.9 & 75.5 & 72.2 & 73.1 & 79.9 & 76.4 & 76.2 & 84.6 & 82.6 & 76.3 & 83.3 & 80.5 & 77.9 & 86.4 & 85.8 \\
    & \checkmark & \checkmark & 82.4 & 91.5 & 91.3 & 83.2 & 91.2 & 90.3 & 83.4 & 92.9 & 92.8 & 83.1 & 90.1 & 88.0 & 83.3 & 92.2 & 92.1 & 83.2 & 92.0 & 92.5 \\
\midrule
\multirow{3}{*}{ViT-B/32~\cite{radford2021learning}}
    & \checkmark & $\times$ & 78.6 & 87.2 & 86.6 & 79.7 & 88.2 & 87.1 & 78.1 & 86.7 & 85.8 & 80.1 & 89.0 & 88.4 & 80.5 & 89.7 & 90.1 & 80.8 & 90.1 & 90.3 \\
    & $\times$ & \checkmark & 59.1 & 61.1 & 62.7 & 63.2 & 66.6 & 66.0 & 63.3 & 67.5 & 67.8 & 65.6 & 71.0 & 68.2 & 68.4 & 73.7 & 70.0 & 67.6 & 71.4 & 65.4 \\
    & \checkmark & \checkmark & 79.1 & 87.4 & 87.4 & 80.2 & 88.2 & 86.6 & 80.1 & 88.4 & 86.9 & 81.3 & 90.4 & 90.4 & 81.7 & 90.6 & 90.8 & 80.3 & 91.2 & 92.1 \\
\midrule
\multirow{3}{*}{ViT-L/14~\cite{radford2021learning}}
    & \checkmark & $\times$ & 87.4 & 95.9 & 96.0 &  87.6 & 96.3 & 96.4 & \underline{92.0} & \underline{97.5} & \underline{97.8} & 89.7 & 97.1 & 97.4 & 86.9 & 95.6 & 95.7 & 86.6 & 96.6 & 97.1 \\
    & $\times$ & \checkmark & 71.5 & 78.6 & 78.0 & 77.1 & 84.5 & 81.8 & 78.5 & 87.3 & 85.9 & 81.4 & 89.6 & 89.9 & 78.5 & 87.3 & 85.9 & 82.8 & 91.4 & 92.0 \\
    & \checkmark & \checkmark & 89.9 & 96.9 & 96.8 &  91.3 & 97.4 & 97.7 & \textbf{92.9} & \textbf{97.9} & \textbf{98.1} & 85.4 & 95.1 & 94.8 & 88.0 & 96.4 & 96.6 & 89.7 & 97.1 & 97.5 \\
\midrule
\bottomrule
\end{tabular}
\end{adjustbox}
\label{tab:diversity}
\vspace{-5pt}
\end{table*}

\begin{table*}[t]
\centering
\caption{Performance comparisons of the proposed approach with SoTA methods across datasets.}
\vspace{-8pt}
\begin{adjustbox}{max width=\textwidth}
\begin{tabular}{c|ccc|ccc|ccc|ccc|ccc|ccc|ccc|ccc}
\toprule
\midrule
\multirow{2}{*}{\textbf{Method}} & \multicolumn{3}{c|}{\textbf{MGD~\cite{wang2020cnn}}} & \multicolumn{3}{c|}{\textbf{CNNDF~\cite{wang2020cnn}}} & \multicolumn{3}{c|}{\textbf{GANDF~\cite{tan2024rethinking}}} & \multicolumn{3}{c|}{\textbf{GENI~\cite{zhu2023genimage}}} & \multicolumn{3}{c|}{\textbf{DIRE~\cite{wang2023dire}}} & \multicolumn{3}{c|}{\textbf{UDF~\cite{ojha2023towards}}} & \multicolumn{3}{c|}{\textbf{MNW~\cite{MNW2025}}} & \multicolumn{3}{c}{\textbf{AVG}} \\
\cmidrule(lr){2-4} 
\cmidrule(lr){5-7} 
\cmidrule(lr){8-10} 
\cmidrule(lr){11-13} 
\cmidrule(lr){14-16}  
\cmidrule(lr){17-19} 
\cmidrule(lr){20-22} 
\cmidrule(lr){23-25}
 & ACC & AUC & AP & ACC & AUC & AP & ACC & AUC & AP & ACC & AUC & AP & ACC & AUC & AP & ACC & AUC & AP & ACC & AUC & AP & ACC & AUC & AP \\
\midrule
AlgnF~\cite{rajan2024aligned} & 64.7 & 69.5 & 73.2 & 67.1 & 73.0 & 74.7 & 66.9 & 72.6 & 73.0 & 80.2 & 86.8 & 90.5 & 79.5 & 86.1 & 91.6 & 75.9 & 83.9 & 87.6 & 45.9 & 43.6 & 44.1 & 68.6 & 73.6 & 76.4 \\
C2PClip~\cite{c2pclip2025} & 94.8 & 97.8 & 98.2 & 95.8 & 99.2 & 99.4 & 95.8 & 99.4 & 99.4 & 76.5 & 94.3 & 95.5 & 73.3 & 94.4 & 96.3 & 92.1 & 99.2 & 99.3 & 43.4 & 56.2 & 48.3 & 81.8 & 91.6 & 91.0 \\
CNND~\cite{wang2020cnn} & 79.9 & 90.2 & 91.0 & 82.9 & 96.5 & 96.3 & 73.4 & 91.8 & 91.7 & 53.7 & 67.5 & 64.8 & 48.8 & 71.2 & 75.8 & 58.3 & 75.8 & 76.7 & 41.3 & 9.9 & 33.6 & 62.6 & 71.8 & 75.7 \\
DMD~\cite{corvi2023detection} & 89.3 & 98.5 & 98.6 & 93.1 & 99.6 & 99.6 & 90.1 & 98.6 & 98.8 & 56.1 & 84.0 & 82.3 & 48.5 & 84.1 & 85.6 & 59.0 & 89.6 & 88.3 & 42.2 & 20.1 & 35.2 & 68.3 & 82.1 & 84.1 \\
FatF~\cite{liu2024forgery} & 89.9 & 98.3 & 98.6 & 98.5 & 99.7 & 99.7 & 87.8 & 99.7 & 99.7 & 81.8 & 93.6 & 95.0 & 80.3 & 95.2 & 97.1 & 93.8 & 98.8 & 99.1 & 43.4 & 11.7 & 38.6 & 82.2 & 85.3 & 89.7 \\
ForL~\cite{chen2025forgelensdataefficientforgeryfocus} & 95.3 & 98.1 & 98.6 & 96.0 & 98.6 & 98.9 & 89.2 & 99.3 & 99.4 & 85.7 & 95.8 & 96.9 & 83.1 & 95.5 & 97.5 & 95.4 & 99.1 & 99.3 & 43.8 & 17.1 & 40.3 & 84.1 & 86.2 & 90.1 \\
FrqNet~\cite{tan2024frequency} & 81.7 & 90.0 & 86.6 & 91.4 & 97.9 & 97.8 & 83.8 & 97.5 & 97.4 & 78.7 & 86.3 & 84.9 & 82.9 & 94.8 & 96.5 & 88.8 & 94.8 & 95.8 & 40.1 & 3.6 & 32.3 & 78.2 & 80.7 & 84.5 \\
LGrad~\cite{tan2023learning} & 45.9 & 47.2 & 45.4 & 85.4 & 92.5 & 91.0 & 56.7 & 73.3 & 71.2 & 62.1 & 68.8 & 67.4 & 83.1 & 95.1 & 96.3 & 83.1 & 89.3 & 90.3 & 58.7 & 44.3 & 63.9 & 67.9 & 72.9 & 75.1 \\
NPR~\cite{tan2024rethinking} & 82.7 & 83.9 & 75.8 & 92.8 & 96.6 & 94.5 & 75.4 & 93.9 & 91.6 & 87.4 & 94.2 & 92.9 & 93.8 & 99.5 & 99.6 & 95.3 & 98.9 & 98.7 & 41.4 & 5.5 & 36.7 & 81.3 & 81.8 & 84.3 \\
RCLip~\cite{cozzolino2024raising} & 55.7 & 86.0 & 85.8 & 80.3 & 88.3 & 87.3 & 79.6 & 93.1 & 92.0 & 86.3 & 92.4 & 94.0 & 86.8 & 93.1 & 95.9 & 89.5 & 94.5 & 96.4 & 60.7 & 76.3 & 63.0 & 79.8 & 89.1 & 87.8 \\
RINE~\cite{koutlis2024leveraging} & 92.8 & 98.2 & 98.4 & 94.6 & 99.5 & 99.6 & 97.3 & 99.7 & 99.7 & 78.2 & 96.6 & 96.7 & 74.4 & 96.1 & 97.0 & 91.1 & 98.8 & 99.0 & 45.4 & 21.9 & 41.9 & 82.0 & 87.3 & 90.3 \\
StayP~\cite{rajan2025stay} & 68.3 & 77.0 & 80.5 & 67.3 & 74.5 & 77.3 & 65.9 & 72.3 & 72.3 & 86.2 & 95.4 & 96.4 & 79.1 & 85.9 & 91.3 & 75.8 & 83.9 & 86.9 & 45.7 & 43.9 & 42.4 & 69.8 & 76.1 & 78.2 \\
UniD~\cite{ojha2023towards} & 80.2 & 94.1 & 94.8 & 87.1 & 97.2 & 97.5 & 89.9 & 96.3 & 96.8 & 66.3 & 87.0 & 87.0 & 62.2 & 88.8 & 90.4 & 82.2 & 95.4 & 96.0 & 44.8 & 30.5 & 41.0 & 73.2 & 84.2 & 86.2 \\
\midrule
$f_p$ & \textbf{92.0} & \textbf{97.5} & \textbf{97.8} & \textbf{95.1} & \textbf{99.2} & \textbf{99.3} & \textbf{96.8} & \textbf{99.8} & \textbf{99.8} & \textbf{83.8} & \textbf{97.3} & \textbf{97.2} & \textbf{85.8} & \textbf{96.3} & \textbf{97.2} & \textbf{91.0} & \textbf{97.5} & \textbf{97.3} & \textbf{80.6} & \textbf{97.5} & \textbf{97.7} & \textbf{89.3} & \textbf{97.9} & \textbf{98.0} \\
$f_p + f_s$ & \textbf{92.9} & \textbf{97.9} & \textbf{98.1} & \textbf{96.0} & \textbf{99.2} & \textbf{99.3} & \textbf{93.1} & \textbf{99.7} & \textbf{99.7} & \textbf{86.0} & \textbf{98.0} & \textbf{97.9} & \textbf{86.6} & \textbf{95.5} & \textbf{96.7} & \textbf{92.6} & \textbf{98.1} & \textbf{98.2} & \textbf{85.1} & \textbf{98.0} & \textbf{98.1} & \textbf{90.3} & \textbf{98.1} & \textbf{98.3} \\
\midrule
\bottomrule
\end{tabular}
\end{adjustbox}
\label{tab:comparisons}
\vspace{-15pt}
\end{table*}

\subsection{Experimental Setup}\vspace{-4pt}
All experiments were conducted on Ubuntu 22.04 using eight NVIDIA RTX 6000 GPUs with 48 GB VRAM each. The proposed dual-branch model was implemented in Python using PyTorch and trained for 10 epochs with a batch size of 128 and a learning rate of 0.0001.\\
We collected nine benchmark datasets commonly used to evaluate generalization, summarized in Table~\ref{tab:dataset_summary}. Each dataset is described by its release year, number of generative models, and the number of real and fake samples. To improve generalization to unseen GAN and diffusion models, different dataset combinations were used. The training set was balanced by sampling equal numbers of real and fake images per generative model, starting with 200 samples and progressively increasing the size, as shown in Table~\ref{tab:diversity}.\\
For evaluation, we used the complete datasets. MGD (mixed GAN and diffusion)~\cite{wang2020cnn, ojha2023towards}, a mixed benchmark containing 12 GAN-based and low-level vision algorithms and 8 diffusion models, was used for comparisons with SoTA methods. Since MNW~\cite{MNW2025} contains only fake samples, an equal number of real samples from ImageNet~\cite{wang2020cnn, ojha2023towards} were used for fair evaluation. Duplicate samples across datasets were identified and removed. \vspace{-5pt}
\subsection{Performance Evaluation and Comparisons}\vspace{-5pt}
This section presents and discusses the results of the proposed data diversity and AIGI detection framework.\vspace{-3pt}
\subsubsection{Performance of Dataset Diversification}
We analyze the impact of training dataset diversity and backbone selection on generalization across traditional and foundation models, as provided in Table~\ref{tab:diversity}. The evaluation is performed on the MGD~\cite{wang2020cnn} benchmark, which comprises both GAN- and diffusion-based generative models.\\
Among traditional CNNs, VGG-16 and ResNet-18 perform close to random guessing across all sample sizes, regardless of feature combinations, highlighting their limited capacity to generalize to unseen GAN and diffusion models. MobileNet-V3~\cite{howard2019searching} and EfficientNet-B4~\cite{tan2019efficientnet} show noticeable improvements with increased data, especially when both feature streams are used, but their performance remains significantly below that of foundation models.\\
In contrast, foundation models demonstrate strong generalization even with fewer training samples. DINOv2~\cite{oquab2023dinov2} and CLIP ViT~\cite{radford2021learning} models consistently outperform traditional CNNs across all data scales. Notably, ViT-L/14 achieves the best performance, reaching an AUC of 97.9 and AP of 98.1 with 19K samples, and maintaining high performance even as the sample size increases or decreases further.\\
Using both feature types jointly ($f_p + f_s$) consistently outperforms using either feature alone, particularly for foundation models, confirming the complementary nature of the proposed dual-branch representation. Overall, these results demonstrate that data diversity combined with foundation models is critical for generalized and robust AIGI detection, while simply increasing training data is insufficient for traditional architectures.
\vspace{-5pt}

\subsubsection{Performance Comparisons}\vspace{-4pt}
The proposed method is compared with existing methods across seven benchmark datasets in Table~\ref{tab:comparisons}. Although several existing methods achieve strong performance on specific datasets, such as MGD~\cite{wang2020cnn, ojha2023towards}, CNNDF~\cite{wang2020cnn}, and GANDF~\cite{tan2024rethinking}, their performance remains inconsistent and degrades significantly on more challenging benchmarks like MNW~\cite{MNW2025}, indicating limited generalization ability.\\
Most SoTA methods experience significant performance degradation on MNW~\cite{MNW2025}, frequently approaching random prediction, whereas the proposed method maintains a substantial margin, achieving 85.1 ACC, 98.0 AUC, and 98.1 AP using the combined features ($f_p + f_s$). This highlights its generalization to difficult, unseen scenarios.\\
On other datasets such as GENI~\cite{zhu2023genimage}, DIRE~\cite{wang2023dire}, and UDF~\cite{ojha2023towards}, the proposed method also consistently outperforms or matches the best SoTA methods, demonstrating stable behavior across both GAN- and diffusion-based datasets. In terms of average performance, it ranks first across all metrics, achieving an average AUC of 98.1 and AP of 98.3, surpassing strong CLIP-based and frequency-domain baselines such as C2PClip~\cite{c2pclip2025}, ForL~\cite{chen2025forgelensdataefficientforgeryfocus}, and RINE~\cite{koutlis2024leveraging}. These results confirm that the proposed data diversity strategy and dual-branch representation provide robust and consistent generalization across heterogeneous generative models, unlike existing approaches whose performance varies widely across datasets. \vspace{-4pt}
\begin{table}[!t]
\centering
\caption{Robustness against Common Perturbations.}
\vspace{-8pt}
\begin{adjustbox}{width=0.48\textwidth}
\begin{tabular}{c c|ccc|ccc|ccc}
\toprule
\midrule
\multirow{2}{*}{\textbf{Method}} & \multirow{2}{*}{\textbf{Purt.}} & \multicolumn{3}{c|}{\textbf{CNNDF~\cite{wang2020cnn}}} & \multicolumn{3}{c|}{\textbf{UDF~\cite{ojha2023towards}}} & \multicolumn{3}{c}{\textbf{MNW~\cite{MNW2025}}} \\
\cmidrule(lr){3-5} \cmidrule(lr){6-8} \cmidrule(lr){9-11}
 & & ACC & AUC & AP & ACC & AUC & AP & ACC & AUC & AP \\
\midrule
\multirow{3}{*}{$f_p$} 
 & Both & 71.2 & 89.0 & 88.1 & 66.7 & 79.4 & 79.6 & 60.7 & 78.8 & 78.3 \\
 & Gaussian & 81.1 & 93.4 & 92.7 & 76.9 & 86.9 & 87.1 & 68.0 & 85.3 & 85.4 \\
 & JPEG & 81.7 & 95.3 & 94.8 & 77.0 & 89.2 & 89.6 & 67.1 & 87.9 & 87.7 \\
 \midrule
\multirow{3}{*}{$f_p$ + $f_s$} 
 & Both & 76.4 & 88.8 & 88.0 & 73.4 & 83.5 & 83.3 & 67.2 & 79.9 & 79.2 \\
 & Gaussian & 82.9 & 92.9 & 92.1 & 79.6 & 88.9 & 88.4 & 73.0 & 85.2 & 85.5 \\
 & JPEG & 85.8 & 95.7 & 95.2g & 83.5 & 93.0 & 92.7 & 74.5 & 89.4 & 88.9 \\
 \midrule
\bottomrule
\end{tabular}
\end{adjustbox}
\label{tab:robustness}
\vspace{-15pt}
\end{table}
\subsubsection{Robustness Analysis}\vspace{-4pt}
To evaluate the robustness of the proposed method, we adopted two commonly used image perturbations: Gaussian blurring and JPEG compression. For training, 50\% of the samples were augmented with these perturbations, using Gaussian blurring with $\sigma$ values of 0.5 and 1.0, and JPEG compression with quality factors of 50 and 70. During testing, the perturbation parameters were randomly selected, with Gaussian blurring $\sigma$ in the range 0--3.0 and JPEG compression quality in the range 30--100, to simulate realistic variations and assess the model's resilience.\\
Table~\ref{tab:robustness} shows that the $f_p + f_s$ method achieves the highest performance across nearly all scenarios. For instance, under Gaussian perturbation, $f_p + f_s$ reaches ACC values of 82.9, 79.6, and 73.0, AUC values of 92.9, 88.9, and 85.2, and AP values of 92.1, 88.4, and 85.5 for CNNDF~\cite{wang2020cnn}, UDF~\cite{ojha2023towards}, and MNW~\cite{MNW2025}, respectively, outperforming $f_p$ alone (ACC: 81.1, 76.9, 68.0; AUC: 93.4, 86.9, 85.3; AP: 92.7, 87.1, 85.4). Under both perturbations, $f_p + f_s$ still maintains higher robustness (ACC: 76.4, 73.4, 67.2; AUC: 88.8, 83.5, 79.9; AP: 88.0, 83.3, 79.2) compared to $f_p$ (ACC: 71.2, 66.7, 60.7; AUC: 89.0, 79.4, 78.8; AP: 88.1, 79.6, 78.3). Overall, Gaussian perturbations yield the highest metrics, JPEG slightly lower, and both consistently produce the lowest performance, indicating that simultaneous noise and compression present the greatest challenge. Across datasets, CNNDF~\cite{wang2020cnn} generally exhibits the highest metrics, UDF~\cite{ojha2023towards} is slightly lower, and MNW~\cite{MNW2025} shows the lowest ACC and AP, highlighting the influence of dataset characteristics. These results clearly demonstrate the advantage of combining features ($f_p + f_s$) to enhance robustness against perturbations.

\begin{table}[!t]
\centering
\caption{Performance across CNNDF~\cite{wang2020cnn}, UDF~\cite{ojha2023towards}, and MNW~\cite{MNW2025} datasets for different training combinations.}
\vspace{-8pt}
\begin{adjustbox}{width=0.48\textwidth}
\begin{tabular}{c|ccc|ccc|ccc}
\toprule
\midrule
\multirow{2}{*}{\textbf{Training Dataset}}
& \multicolumn{3}{c|}{\textbf{CNNDF~\cite{wang2020cnn}}} 
& \multicolumn{3}{c|}{\textbf{UDF~\cite{ojha2023towards}}} 
& \multicolumn{3}{c}{\textbf{MNW~\cite{MNW2025}}} \\
\cmidrule(lr){2-4}
\cmidrule(lr){5-7}
\cmidrule(lr){8-10}
& ACC & AUC & AP 
& ACC & AUC & AP 
& ACC & AUC & AP \\
\midrule
PGAN & 94.2 & 99.2 & 99.2 & 91.0 & 94.3 & 94.3 & 65.8 & 75.2 & 74.8 \\
D1KS & 90.4 & 98.1 & 98.2 & 81.6 & 92.1 & 92.3 & 74.4 & 80.5 & 81.2 \\
DIRE & 69.5 & 91.7 & 87.3 & 88.3 & 97.2 & 95.0 & 71.1 & 77.1 & 76.5 \\
GENI & 91.5 & 97.9 & 98.0 & 92.6 & 95.9 & 96.9 & 67.4 & 73.5 & 74.1 \\
\midrule
PGAN+GENI & 94.6 & 98.0 & 98.3 & 92.4 & 95.9 & 95.9 & 72.7 & 78.8 & 79.0 \\
PGAN+DIRE & 94.2 & 99.2 & 99.2 & 85.0 & 90.3 & 92.3 & 75.8 & 81.5 & 81.7 \\
\textbf{PGAN+D1KS} & \textbf{96.0} & \textbf{99.2} & \textbf{99.3} & \textbf{92.6} & \textbf{98.1} & \textbf{98.2} & \textbf{85.1} & \textbf{98.0} & \textbf{98.1} \\
\midrule
PGAN+D1KS+GENI & 95.1 & 98.7 & 99.0 & 88.4 & 94.1 & 94.8 & 78.5 & 86.2 & 87.6 \\
PGAN+D1KS+DIRE & 93.6 & 98.4 & 98.5 & 90.2 & 94.8 & 94.3 & 77.9 & 85.3 & 84.9 \\
\midrule
\bottomrule
\end{tabular}
\end{adjustbox}
\label{tab:combinations}
\vspace{-15pt}
\end{table}

\begin{figure*}[!t]
    \centering
    \subfloat[\label{fig:ROC_SelfSynthesisGAN_AttGAN}]{{\includegraphics[width=0.68\columnwidth]{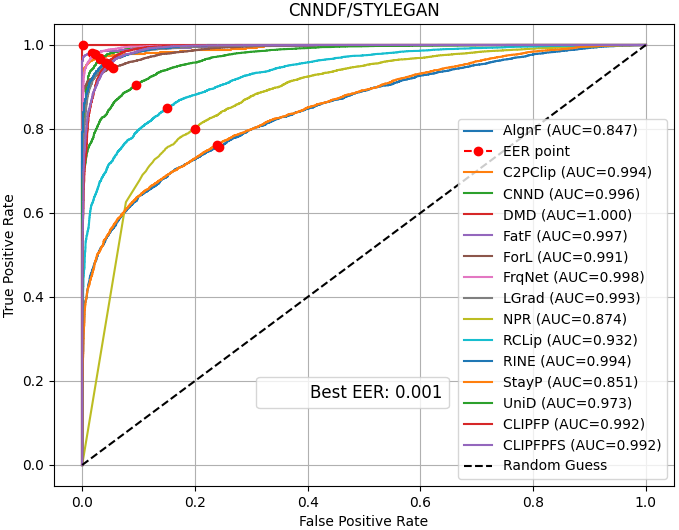}}}
    \subfloat[\label{fig:ROC_GENI_DALLE2}]{{\includegraphics[width=0.68\columnwidth]{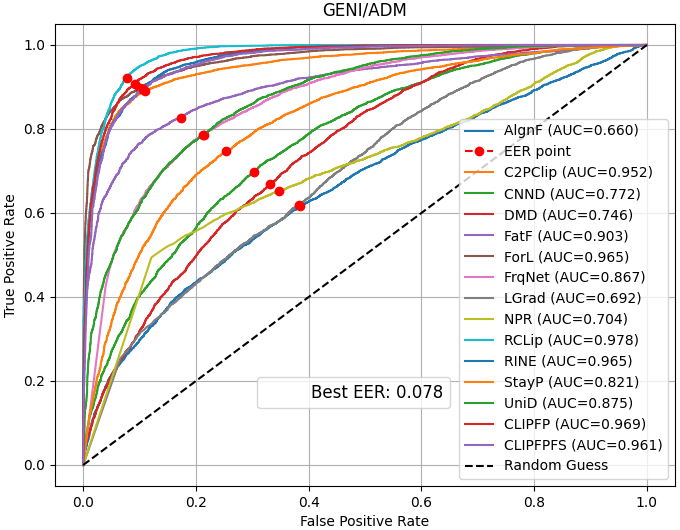}}}
    \subfloat[\label{fig:ROC_ForenSynthsCh_crn}]{{\includegraphics[width=0.68\columnwidth]{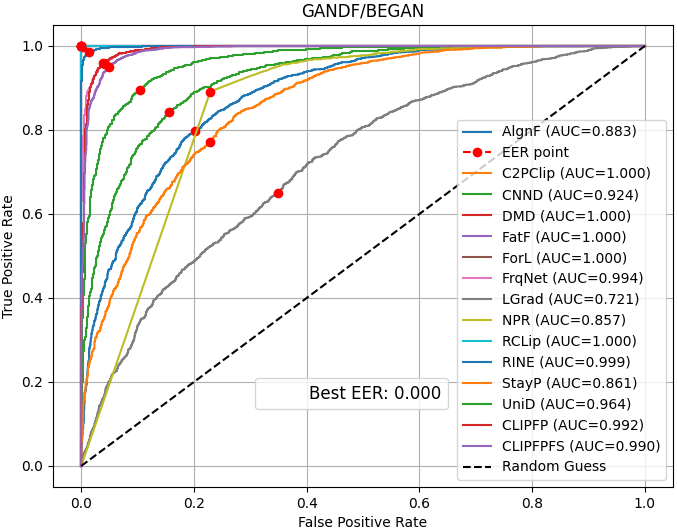}}}
    \vspace{-8pt}
    \caption{Comparison of ROC curves for SoTA and the proposed AIGI detection models, with AUC and EER reported. Panels correspond to: (a) CNNDF/StyleGAN~\cite{wang2020cnn}, (b) GENI/ADM~\cite{zhu2023genimage}, and (c) GANDF/BEGAN~\cite{tan2024rethinking}. AUC indicates overall discrimination, whereas EER measures balanced false acceptance and false rejection rates.}
    \label{fig:roc_curve}
    \vspace{-15pt}
\end{figure*}

\subsection{Ablation Study}\vspace{-4pt}
This section presents an ablation study with dataset configurations and ROC analysis to explain the model behavior. \vspace{-14pt}
\subsubsection{Training Dataset Configurations}\vspace{-4pt}
Table~\ref{tab:combinations} presents the performance of the proposed model when trained with different combinations of datasets and evaluated on CNNDF~\cite{wang2020cnn}, UDF~\cite{ojha2023towards}, and MNW~\cite{MNW2025}, respectively. Individual datasets such as PGAN~\cite{karras2017progressive}, D1KS~\cite{tan2024rethinking}, DIRE~\cite{wang2023dire}, and GENI~\cite{zhu2023genimage} show varying generalization capabilities, with MNW~\cite{MNW2025} consistently being the most challenging.\\
Combining datasets generally improves performance, especially on MNW~\cite{MNW2025}, highlighting the benefits of data diversity. Notably, the PGAN+D1KS combination achieves the best overall results, reaching 85.1 ACC, 98.0 AUC, and 98.1 AP on MNW~\cite{MNW2025}, while maintaining strong performance on CNNDF~\cite{wang2020cnn} and UDF~\cite{ojha2023towards}. Adding additional datasets (e.g., GENI~\cite{zhu2023genimage} or DIRE) does not further improve MNW~\cite{MNW2025} performance, indicating potential catastrophic forgetting and suggesting that a carefully selected set of complementary datasets is sufficient to achieve generalization. Overall, these results demonstrate that training with diverse yet complementary datasets significantly enhances model robustness across challenging unseen distributions.\vspace{-4pt}
\subsubsection{Analysis of ROC Curves}\vspace{-4pt}
The ROC curves in Figure~\ref{fig:roc_curve} illustrate the detection performance of the proposed method compared to several SoTA approaches across six sub-datasets, such as CNNDF/StyleGAN~\cite{wang2020cnn}, GENI/ADM~\cite{zhu2023genimage}, and GANDF/BEGAN~\cite{tan2024rethinking}. The proposed method (CLIP with pixel feature-CLIPFP and CLIP with pixel and spectral features-CLIPFPFS) consistently achieves curves near the top-left corner, indicating high true positive rates with low false positives, and attains the lowest equal error rates (EER) in most cases. While some SoTA methods, such as C2PClip~\cite{c2pclip2025}, RINE~\cite{koutlis2024leveraging}, and ForL~\cite{chen2025forgelensdataefficientforgeryfocus}, perform well on specific datasets, their performance drops significantly on more challenging scenarios like GENI/ADM~\cite{zhu2023genimage} and GANDF/BEGAN~\cite{tan2024rethinking}. In contrast, the proposed method maintains high AUC and low EER across all datasets, demonstrating better generalization across both GAN- and diffusion-based generative models.\vspace{-5pt}
\subsection{Discussions, Limitations, and Future Insights}\vspace{-4pt}
This section provides several critical insights into the relationship between training dataset diversity and generalizability, as well as limitations and future insights.
\vspace{-8pt}
\subsubsection{Discussions}\vspace{-4pt}
\textbf{Redundancy Trap in Standard Datasets: }
Existing AIGI detectors trained on unfiltered dense datasets exhibit poor generalization across unseen generative models, indicating memorization of redundant patterns rather than learning transferable features. In contrast, the proposed feature-space filtering forms sparse, diverse clusters that promote generalized artifact learning and mitigate overfitting. \\
\textbf{Role of Sample Diversity over Quantity:}
Performance does not scale linearly with data volume. For example, ViT-L/14 achieves 98.1\% AP using only 19K filtered samples (T=0.8), while increasing to 180K unfiltered samples (T=1.0) yields AP of 97.5\%. This demonstrates that diversity-aware deduplication is more critical for cross-model generalization than data quantity.\\
\textbf{Effectiveness of Dual-Branch Design:}
The fusion of pixel- and frequency-domain representations is crucial for modeling the heterogeneous artifacts introduced by modern generative models. The pixel-domain branch captures high-level semantic inconsistencies, while the frequency-domain branch exposes periodic patterns and structural irregularities that are less visible in the spatial domain. Experimental results confirm that their fusion through a trainable detection head substantially improves the detection of unseen synthetic artifacts compared to single-branch designs.\\
\textbf{Robust Cross-Paradigm Generalization:}
Unlike single-source trained models that fail across generative paradigms, Diversity Matters maintains strong performance over 12 GAN and 8 diffusion models. Inter- and intra-class filtering narrows decision boundary gaps, substantially improving cross-dataset robustness.
\vspace{-4pt}
\subsubsection{Limitations}\vspace{-4pt}
While the proposed approach improves generalization and robustness, it relies on pre-defined perturbations and may be less effective against unseen~\cite{MNW2025} or more complex attacks~\cite{uddin2019anti, uddin2021analysis}. Additionally, MNW~\cite{MNW2025} requires sampling real images to balance evaluation, which may not fully capture real-world variability.\vspace{-4pt}

\subsubsection{Future Insights}\vspace{-4pt}
Future work should explore adaptive feature fusion and meta-learning strategies to generalize across unseen perturbations and generative models~\cite{uddin2025adversarial, uddin2025advbench}. Investigating lightweight~\cite{howard2019searching}, real-time implementations~\cite{uddin2025guard}, and text-visual/3D representations~\cite{uddin2022deep} cues could further enhance robustness and generalization for practical systems. A continual learning~\cite{wang2026generalizable} mechanism that dynamically updates parameters with data can further improve adaptability to previously unseen zero-day attack distributions.
\vspace{-8pt}
\section{Conclusion}\vspace{-5pt}
In this work, we proposed a novel framework for AIGI detection that emphasizes both data diversity and complementary feature domains. By filtering redundant samples and combining pixel- and frequency-domain CLIP features, the proposed method achieves improved generalization across unseen generative models, datasets, and adversarial conditions. Extensive experiments demonstrate that enhancing dataset diversity and leveraging complementary feature representations are crucial for building robust, generalized, and reliable detectors in the rapidly evolving landscape of synthetic content. Future work will explore further generalization of the proposed method to novel generative techniques and more challenging real-world scenarios, including the continuous emergence of newly designed generative and adversarial models.

{
    \small
    \bibliographystyle{ieeenat_fullname}
    \bibliography{main}
}


\end{document}